# Advancing Financial Risk Prediction Through Optimized LSTM Model Performance and Comparative Analysis

Ke Xu[1], Yu Cheng[2], Shiqing Long[3], Junjie Guo[4], Jue Xiao[5], Mengfang Sun[6]

[1]Columbia University,USA,Katiexu.kx@gmail.com

[2]Columbia University,USA,yucheng576@gmail.com

[3]Independent Researcher,China,krystenlong0606@gmail.com

[4]Rutgers University,USA,jg1806@scarletmail.rutgers.edu

[5]University of Connecticut Business School,USA,juexiaowork@gmail.com

[6]Stevens Institute of Technological,USA,smengfang@gmail.com

*This paper focuses on the application and optimization of LSTM model in financial risk prediction. The study starts with an overview of the architecture and algorithm foundation of LSTM, and then details the model training process and hyperparameter tuning strategy, and adjusts network parameters through experiments to improve performance. Comparative experiments show that the optimized LSTM model shows significant advantages in AUC index compared with random forest, BP neural network and XGBoost, which verifies its efficiency and practicability in the field of financial risk prediction, especially its ability to deal with complex time series data, which lays a solid foundation for the application of the model in the actual production environment.*

## I. INTRODUCTION

Risks in finance encompass credit, market, and operational aspects, which are the core issues that financial institutions must face in their daily operations [1]. Traditional risk prediction models mostly rely on statistical methods and simple machine learning algorithms. Although these methods can identify risk factors to a certain extent, they have limitations in dealing with high-dimensional data, capturing nonlinear relationships, and dealing with data noise. With the surge of financial trading volume and the improvement of data complexity, the shortcomings of traditional models have become increasingly prominent, and it is difficult to meet the high requirements of modern financial markets for risk prediction accuracy and speed [2].

Through the multi-layer neural network structure, deep learning can automatically learn complex feature representations from massive data, which can not only effectively deal with high-dimensional data, but also capture the deep nonlinear relationship between variables, which is particularly important for understanding the complex dynamics of financial markets [3]. In addition, the deep learning model also has good generalization ability, which can effectively predict risk on the basis of limited samples and reduce the risk of overfitting [4]. Coupled with its powerful parallel computing capabilities, the risk analysis process is more rapid, which helps financial institutions to quickly respond to market changes and adjust risk management strategies in time.

In today's era of accelerated digital transformation, the financial industry is undergoing unprecedented changes. With the rapid development of big data technology and the continuous breakthrough in the field of artificial intelligence, deep learning, as an important branch of artificial intelligence, has gradually penetrated into all levels of financial risk management and become one of the key tools for predicting and controlling financial risks [5]. This paper aims to explore how to use deep learning technology to improve traditional risk assessment methods, introduce more efficient and accurate analysis tools for the field of financial risk prediction, improve the accuracy and efficiency of financial risk prediction, and provide more scientific and timely risk management strategies for financial institutions.

## II. CORRELATIONAL RESEARCH

As artificial intelligence technology progresses swiftly, the output of academic literature has risen significantly. Deep learning has made remarkable achievements in the fields of medical diagnosis classification[6-9], computer vision[10-13], and disease prediction[14]. In addition, stock market trend prediction, financial risk quantification, and asset allocation optimization have become key issues to be solved urgently. In this wave of technology, deep learning, as a cutting-edge branch of machine learning, is leading the innovation. Compared with previous machine learning models, deep learning does not require manual preprocessing and feature selection steps. On the contrary, it relies on a multi-level nonlinear structure to automatically perform feature extraction and transformation. These hierarchical structures enable the network to capture the inherent complex nonlinear correlations of data, and directly extract the learned high-level features from the original input [15], which greatly enhances the model's ability to recognize and understand data patterns.

Zhou et al. [16] showed that on the Hadoop cloud computing platform, the deep learning model integrated with Convolutional Neural Network (CNN) can quickly and accurately identify fraudulent transactions in supply chain finance, emphasizing that the integration of cloud computing and deep learning in dealing with financial risk identification, especially in the field of supply chain, is a highly potential development direction. Stevenson et al. [17] used BERT, the natural language processing technology of deep learning, to mine rich information from the text of corporate financial reports, which helped to alleviate the information asymmetry problem of SME financing. Facing the challenge of sample imbalance in credit default risk identification, Lin et al. [18] successfully used Generative Adversarial Networks (GAN) to improve the accuracy of credit default swap (CDS) risk assessment. In addition, deep learning has also shown innovative applications in other aspects of financial risk management. For example, Liu et al. [19] developed a video surveillance system specifically for risk review to enhance stock trading compliance by applying deep learning technology to computer vision. The fuzzy deep learning model designed by Pena et al. [20] followed the Basel III guidelines and effectively predicted operational risks.

Adha and Nurrohmah adopted the multinomial logistic regression model to predict the proportion of bank's default loss, and used maximum likelihood estimation to optimize the model parameters. Experimental results show that the accuracy of the model in the classification of bank customer default loss reaches 95.3% [21]. On the other hand, Silvia and Paolo used Bayesian model averaging strategy to construct credit risk model, and proved through empirical analysis that on a real credit risk database, this method not only improved the prediction accuracy, but also showed better comprehensive performance compared with the basic regression model [22].

Yannis and Magdalene used the loan records of 1411 companies from a major commercial bank in Greece to compare and analyze the two classification models of support vector machine and decision tree, and found that support vector machine was slightly better than decision tree in stability [23]. Alina and Simona respectively used Bayesian logistic regression and artificial neural network model for commercial default risk. Based on the sample of 3000 enterprises of a multinational bank in Romania, the study pointed out that artificial neural network was superior to the Bayesian model in terms of prediction accuracy and effect [24].

Zhao Nan, Zhao Zheyun and other researchers combined principal component analysis and BP neural network technology to provide effective ways and practical plans for credit risk early warning for Internet financial institutions [25]. In order to cope with the challenges of financial big data, Yang Dejie and Zhang Ning et al. improved the stack denoising autoencoder neural network model and introduced Karhunen-Loeve expansion as a means of noise processing. Experimental results show that this improvement improves the accuracy of the model by about 3% on the benchmark [26]. In addition, Yi Baiheng and Zhu Jianjun et al., aiming at the problem of unbalanced samples, proposed a method of synthetic data enhancement only for misclassified samples, so as to optimize the classification hyperplane bias problem of support vector machine in dealing with such samples, applied this enhancement to assess customer credit risk for small loan firms, outperforming other methods in accuracy [27].

## III. METHOD

### A. LSTM

LSTM (Long Short-Term Memory network), an advanced class of recurrent neural networks, is engineered to address the challenge of long-term dependencies in sequence data processing. This network employs an innovative gating mechanism comprising the input gate, forget gate, cell state, and output gate, facilitating efficient selective information storage, forgetting, and output. Consequently, LSTMs have proven their superior capabilities in complex network domains such as natural language processing[28], speech recognition[29], time series prediction[30], video analytics[31-33], and material detection[34]. Compared with the traditional RNN model, LSTM relies on sigmoid and tanh functions to fine-regulate the information flow, significantly alleviate the phenomenon of gradient disappearance and explosion[35], and ensure that the model can firmly grasp and use historical information even when dealing with long sequence data to achieve higher accuracy prediction and content generation. LSTM is also frequently integrated with other cutting-edge technologies to continuously expand the frontier of sequence data processing [36]. This study develops a deep learning model featuring an input layer, three intermediate hidden layers, and a final output layer, and its architecture layout is shown in Figure 1 to further explore the potential and application of LSTM in network structure design.

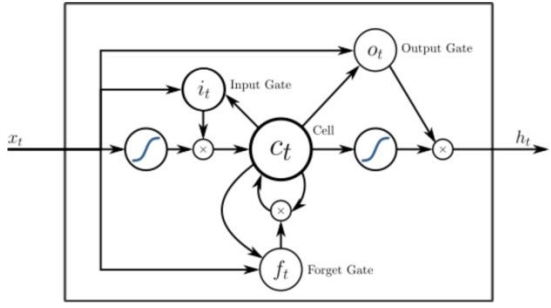

Figure 1. LSTM Long Short-Term Memory network structure

1. Algorithm flow

First, we analyze the forward propagation process of LSTMs, encompassing the entire pathway of information from the input layer to the output layer. During this propagation, the input gate, output gate, and forget gate determine whether to retain or discard data.

The input gate incorporates two sources of information: a weighted sum of the current input vector and another weighted sum from the previous time step's hidden states, as expressed in Equation (1).

$$\chi_\rho^t = 2\Sigma_{i=1}^I w_{i\rho} x_i^t + \Sigma_{h=1}^H w_{h\rho} y_h^{t-1} \quad (1)$$

The output generated by the input gate is calculated by a specific function, and the calculation process follows equation (2).

$$y_p^t = f(x_\rho^t) \quad (2)$$

The hidden layer's memory unit processes information via a weighted combination of the current input vector and the prior hidden state, adhering to Formula (3).

$$\chi_c^t = \Sigma_{i=1}^I w_{ic} x_i^t + \Sigma_{h=1}^H w_{hc} y_h^{t-1} \quad (3)$$

The state feedback value received by the input gate from the memory unit of the hidden layer is reflected in Formula (4).

$$S_c^t = 2s_c^{t-1} + y_\rho^t g(2x_C^t) \quad (4)$$

The input and output formulas of the forget gate are shown in Formula (5) and (6).

$$\chi_\varphi^t = 2^I w_{i\varphi} x_i^t + 2_{h=1}^H w_{h\varphi} y_h^{t-1} \quad (5)$$
$$y_\varphi^t = f(x_\varphi^t) \quad (6)$$

The state feedback received by the forget gate from the memory unit of the hidden layer is calculated according to Formula (7).

$$S_\varphi^t = s_\varphi^{t-1} + y_\rho^t g(x_\varphi^t) \quad (7)$$

The input received by the output gate is derived from the weighted sum of the memory unit and the output layer, following the stipulation of Formula (8).

$$\chi_\pi^t = 2\Sigma_{i=1}^I w_{i\pi} x_i^t + 2\Sigma_{h=1}^H w_{h\pi} y_h^t \quad (8)$$

The final output result of the memory unit of the hidden layer is calculated and determined according to Formula (9).

$$y_c^t = y_\pi^t h(s_c^t) \quad (9)$$

2. VAR model

Vector autoregressive models are a method for analyzing multivariate time series data that captures the dynamic interaction between each variable by regressing it with respect to its historical value and the historical values of other variables. Its model formula is given by Formula (10).

$$Y_t = A_1 Y_{t-1} + A_2 Y_{t-2} + \cdots + A_x Y_{t-x} \quad (10)$$

Here, $Y$ represents the vector form of the endogenous variable, $A$ is the coefficient matrix, and $x$ refers to the lag order of the variable.

One of the characteristics of vector autoregressive (VAR) model is that it is less dependent on economic theory, only based on a few assumptions, focusing on time series data, and taking it as the core to describe the reaction path of economic system to various shocks. By analyzing the shock response, VAR model can reveal the stability characteristics and transmission effects of dynamic changes in the economic system. The model skillfully integrates theoretical and practical data, uses simple linear or nonlinear regression techniques to establish the internal relationship between variables, and constitutes a comprehensive analysis framework composed of multiple equations.

3. β coefficient of financial risk

The Beta coefficient of financial risk is a key indicator to measure the sensitivity of a single asset or portfolio to the overall volatility of the market, which is widely used in the Capital Asset Pricing model (CAPM). It quantifies systematic risk, that is, the risk that diversification cannot eliminate, by analyzing the proportional relationship between the return on an asset and the change in the return on a market index, such as the S&P 500. The β coefficient signifies market responsiveness of assets: a value of 1 implies assets move in tandem with the market; exceeding 1 suggests higher volatility compared to the market; below 1 denotes less volatility, and a rare instance below 0 signifies the asset moves counter to the market. In the practice of risk management, β coefficient is an effective tool to adjust the risk level of the portfolio and help to build a portfolio that conforms to the risk preferences of investors. At the same time, it only measures the systematic risk rather than the non-systematic risk of a specific company or industry. Through statistical regression analysis of historical data, it is calculated that β coefficient not only predicts the expected return rate of assets, but also plays a key parameter to balance risk exposure in hedging strategies, thus playing a core role in financial market analysis, investment decision-making and risk management. The formula for calculating β coefficient is shown in Equation (11).

$$\beta_i = \frac{Cov(R_i, R_m)}{Var(Rm)} \quad (11)$$

In this context, $Cov(R_i, R_m)$ indicates the covariance between the returns of the $i^{th}$ asset and the returns of the market portfolio, while $Var(R_m)$ denotes the variance of the market portfolio's returns.

4. Description of LSTM model parameters

The parameters of deep neural network model are divided into two categories: manually set hyperparameters and self-learning weights of the model. After data input, the model adjusts the weight of each connection in the training process. According to the preset learning rate, LSTM gradually optimizes the weight according to the prediction error until it reaches the preset upper limit of iteration. Hyperparameters, such as network depth, activation function selection (this study uses the typical Sigmoid function in LSTM), number of hidden layer nodes (64, 32, 16 layers in this case), loss function (choose mean square error), optimization algorithm (use efficient adam algorithm), batch size (set to 100), and total iteration rounds (1000). It needs to be defined in advance, which has a direct impact on the model prediction performance. An overview of the LSTM configuration detailed below is presented in Table 1.

TABLE I. LSTM CONFIGURATION

| Layer | Output Shape | Param |
|---|---|---|
| Lstm1 | (None, 1, 64) | 35072 |
| Lstm2 | (None, 1, 32) | 12416 |
| Lstm3 | (None, 1, 16) | 3136 |
| Dense5 | (None, 1) | 17 |
| Activation6 | (None, 1) | 0 |

B. *Parameter tuning and optimization*

1. Evaluation metrics

Classification model evaluation metrics serve as crucial instruments to gauge the predictive power of models, enabling assessment of their accuracy in categorizing unseen data.

1) Accuracy

Accuracy, a primary evaluation metric, represents the proportion of correctly classified instances by the model relative to the total number of instances.

$$\text{Accuracy} = \frac{TP + TN}{TP + TN + FP + FN} \quad (12)$$

TP stands for true positives, TN for true negatives, FP for false positives, and FN for false negatives.

2) Precision and Recall

Precision is a measure of the fraction of examples that the model predicts to be positive are actually positive.

$$\text{Precision} = \frac{TP}{TP + FP} \quad (13)$$

Determine the proportion of true positive instances detected by the model, which represents the fraction of all actual positive instances correctly identified by the model.

$$\text{Recall} = \frac{TP}{TP + FN} \quad (14)$$

3) F Score

F Score, a unified measure of precision and recall, is their harmonic mean.

$$F = \frac{2 \times Precision \times Recall}{Precision + Recall} \quad (15)$$

4) ROC Curve and AUC

The Receiver Operating Characteristic (ROC) curve serves as a graphical tool for assessing binary classifiers, widely utilized in fields such as healthcare, statistics, and machine learning. This curve illustrates the balance between true positive rates (TPR) and false positive rates (FPR), thereby demonstrating the classifier's performance across different threshold levels. The Area Under the Curve (AUC) provides a single metric summarizing the ROC curve's overall effectiveness, with scores ranging from 0 to 1. A score approaching 1 indicates an exceptional model proficient in distinguishing between positive and negative cases. An AUC of 0.5 means that the model is not discriminative, which is equivalent to random guessing. Anything below 0.5 indicates poor model performance. Therefore, ROC curve and AUC value together constitute an important tool to evaluate the balance between classification ability, sensitivity and specificity of the model.

2. Optimizing Network Depth and Neuron Counts per Layer

The LSTM architecture comprises an input, hidden, and output layer. The input layer aligns with the feature count, necessitating 45 nodes to fit the dataset's structure. The output layer mirrors classification targets, hence, catering to the binary outcomes—"approved" or "rejected"—a solitary output node is employed.

1) Layer Count Optimization

In examining network depth, other hyperparameters were held constant: 20 hidden nodes per layer, Sigmoid activation, logarithmic loss function, Adam optimizer, a batch size of 100, and 100 epochs. Four architectural setups were considered: a single hidden layer, two hidden layers, three hidden layers, and four hidden layers. Comparative analysis was conducted using experimental datasets for these configurations, with the resultant test set Loss curves plotted in Figure 2.

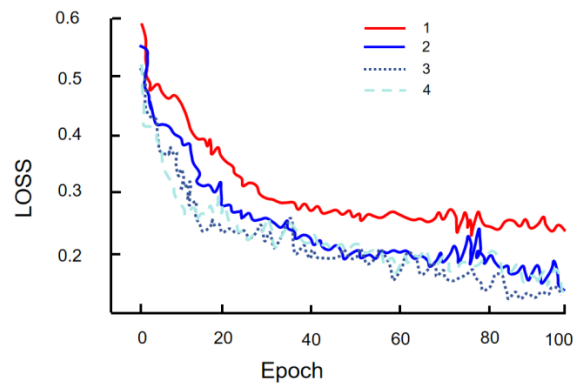

Figure 2. Plot of loss for different number of layers

Figure 2 illustrates that the loss decreases to a minimal point with three network layers, and further layer increments yield negligible loss reduction. Post-training, Table II summarizes the terminal loss and AUC scores of the models.

TABLE II. LOSS WITH DIFFERENT NUMBER OF LAYERS, AUC VALUES

| Network layer number | Training set | | Test set | |
|---|---|---|---|---|
| | LOSS | AUC | LOSS | AUC |
| 1 | 0.2773 | 0.8836 | 0.3394 | 0.5237 |
| 2 | 0.2020 | 0.9125 | 0.2402 | 0.5408 |
| 3 | 0.1758 | 0.9185 | 0.1845 | 0.5384 |
| 4 | 0.1792 | 0.9220 | 0.1994 | 0.5349 |

During the experiment varying hidden node counts, all other hyperparameters remained constant: three hidden layers, Sigmoid activation, logarithmic loss, Adam optimization, a batch size of 100, and 100 iterations. We examined the following four configurations of hidden layer nodes: model1 has 20 nodes per layer (20, 20, 20), model2 has 60 nodes per layer (60, 60, 60), model3 has 100 nodes per layer (100, 100, 100), and model4 uses decreasing number of nodes (64, 32, 16). These four models are tested by applying the data of this study, and their Loss function (Loss) curves on the test set are plotted, and the specific results are shown in Figure 3.

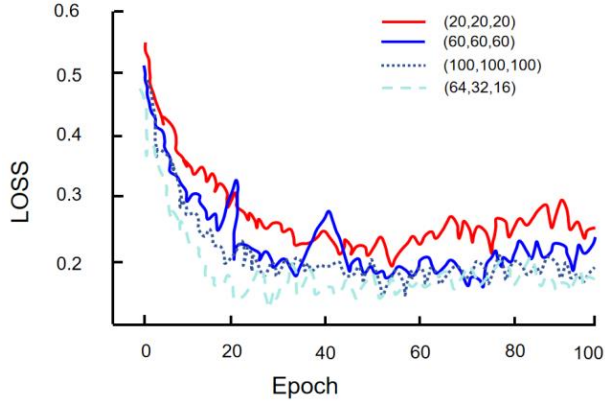

Figure 3. Plot of loss with different number of nodes

Figure 3 indicates that the loss attains a notably low level at node counts of (64, 32, 16), with subsequent increases in nodes showing marginal loss reduction. Upon completing training for all models, their final loss and AUC metrics are summarized in Table III.

TABLE III. LOSS WITH DIFFERENT NUMBER OF NODES, AUC VALUES

| Node number | Training set | | Test set | |
|---|---|---|---|---|
| | LOSS | AUC | LOSS | AUC |
| (20,20,20) | 0.1425 | 0.9315 | 0.2502 | 0.8005 |
| (60,60,60) | 0.1139 | 0.9527 | 0.2833 | 0.8348 |
| (100,100,100) | 0.0923 | 0.9618 | 0.3309 | 0.7964 |
| (64,32,16) | 0.0918 | 0.9577 | 0.2212 | 0.8031 |

The data in Table 3 show that when the hidden layer structure is set to (64, 32, 16) nodes, the AUC index of the model reaches a better level. Therefore, based on the above analysis, a 3-layer hidden layer structure with 64, 32 and 16 nodes per layer was determined as the best configuration for the model in this study.

## IV. APPLICATION OF MODEL

### A. Data set and experimental environment

1. Data set

The experimental Data are collected from the World Bank Open Data dataset, which is a comprehensive Open data resource maintained by the World Bank. To process the experimental data collected from the World Bank Open Data dataset, we utilized a linked data method as proposed in recent literature on enhancing data accessibility and interoperability [37]. This approach allowed us to effectively integrate and analyze the comprehensive range of indicators provided by the World Bank Open Data. The dataset includes over 9,000 indicators spanning various dimensions such as economic, social, environmental, and others, encompassing macroeconomic factors like GDP, inflation, and employment, as well as social progress metrics such as education, health, and poverty. Additionally, it includes data on international trade, foreign direct investment, and environmental impacts. The World Bank Open Data supports multiple languages and data formats, facilitating easy data filtration, search, and visualization through its user-friendly interface. This accessibility is crucial for stakeholders including policymakers, researchers, and NGOs, enabling enhanced data-driven decision making. By applying the linked data method, we were able to leverage the dataset's potential to its fullest, fostering a deeper understanding of global economic phenomena and contributing significantly to the fields of financial risk prediction, market analysis, and sustainable development.

2. Experimental environment

The experimental environment adopted in this paper is shown in Table IV

TABLE IV. EXPERIMENTAL SETUP

| Environmental form | Details |
|---|---|
| CPU | Intel Core i7-9750 |
| GPU | NVIDIA GeForce GTX 1660 Ti |
| Internal storage | 32G |
| Hard disk | 1T SSD |
| Operating system | Linux Ubuntu 16.04 |
| Software | Python 3.7.4 |

### B. Experimental results and analysis

To assess our proposed LSTM deep learning model's efficacy, comparative experiments are conducted against outputs from other prevalent models, including Random Forest[38], XGBoost[39], and traditional Backpropagation Neural Networks.

By fixing all parameters related to pseudo-randomness and implementing five-fold cross-validation strategy, this paper aims to enhance the stability of the algorithm and the reliability of the results. After completing the training of each model, the performance comparison details are listed in Table 5.

TABLE V. COMPARISON OF FINAL RESULTS OF DIFFERENT MODELS

| Model | Acc | Precision | Recall | F | AUC |
|---|---|---|---|---|---|
| SVM | 0.9102 | 0.8345 | 0.7729 | 0.8025 | 0.7924 |
| XGBoost | 0.9373 | 0.9109 | 0.8124 | 0.8588 | 0.8105 |
| BP | 0.8846 | 0.7012 | 0.7283 | 0.7145 | 0.7538 |
| Model of this paper | 0.9731 | 0.8736 | 0.8426 | 0.8578 | 0.8522 |

The data in Table 5 show that the random forest model shows a high accuracy rate of 0.9784, but in view of the sample skew characteristics in the financial risk control scenario, a high accuracy rate is not enough to comprehensively evaluate the performance of the model. It is worth noting that the XGBoost model outperforms in F-measure, while the LSTM model dominates in AUC metric. Although XGBoost is currently widely used in production environments, LSTM shows higher potential and adaptability because it is good at dealing with time-series related and class imbalance data, especially in sequence problems such as financial risk control, implying that LSTM model may have better comprehensive performance.

To further validate LSTM's efficacy as a classification model, Performance-Recall curves and ROC plots are utilized for comparison, illustrated in Figure 4.

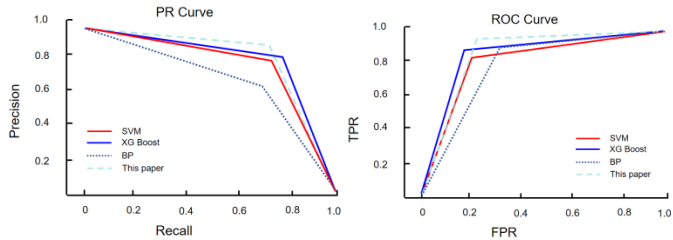

Figure 4. P-R and ROC plots

ROC curve and P-R curve are both effective tools to evaluate the generalization and classification performance of a model on a specific data set. From the analysis of P-R curve, except logistic regression, the other models show good performance, especially the random forest algorithm. However, through ROC curve analysis, the LSTM model shows the superiority of its classification performance with the largest area under the curve (AUC). Considering that in actual deployment, models such as random forest and XGBoost may involve more parameter tuning due to stability requirements, which is time-consuming and costly, compared with LSTM model, the simplicity of parameter tuning becomes a significant advantage. In summary, the LSTM model is an ideal choice for production applications due to its excellent comprehensive performance and easy adjustment.

## V. CONCLUSIONS

This section delves into the specifics of our LSTM-based risk control model implementation. We commence with an outline of the LSTM model's architectural design and fundamental algorithmic tenets, followed by detailing the training procedure and crucial hyperparameter settings. A meticulous experimentation phase ensued, wherein we refined the network configuration by modulating both the depth of hidden layers and the neuron count in each, in pursuit of optimization. The resultant, fine-tuned LSTM model's all-round performance was then benchmarked against alternatives: random forests, backpropagation neural networks, and XGBoost. Empirical findings highlighted the LSTM model's prowess, particularly in achieving high Area Under the Curve (AUC), affirming its fitness and superiority for deployment in real-world operational contexts. Looking forward, the adaptability of LSTM to various financial scenarios suggests promising avenues for further research, particularly in integrating real-time data analysis for more dynamic risk assessment. Future work could explore the integration of additional data sources and the use of ensemble techniques to further enhance predictive accuracy and robustness. Additionally, studies could assess the LSTM model's performance across different financial markets to generalize its applicability and effectiveness. Ultimately, this research contributes to the ongoing evolution of risk management strategies within the financial sector, supporting the shift towards more data-driven, precise, and timely decision-making processes. By continuing to refine these deep learning models, the financial industry can better anticipate and mitigate risks, safeguarding against potential crises and enhancing overall market stability.